\documentclass[10pt,twocolumn,letterpaper]{article}

\usepackage[pagenumbers]{cvpr} % To force page numbers, e.g. for an arXiv version
\usepackage{verbatim} % TO use comment remvoign display of template text
\usepackage{multirow}
\usepackage{booktabs}
\usepackage[table]{xcolor}
\usepackage{adjustbox}

\definecolor{cvprblue}{rgb}{0.21,0.49,0.74}
\usepackage[pagebackref,breaklinks,colorlinks,citecolor=cvprblue]{hyperref}

\def\paperID{8} % *** Enter the Paper ID here
\def\confName{CVPR}
\def\confYear{2024}
\title{Impact of Video Compression Artifacts on \\ Fisheye Camera Visual Perception Tasks}

\author{Madhumitha Sakthi$^{1}$, Louis Kerofsky$^{1}$, Varun Ravi Kumar$^{1}$ and Senthil Yogamani$^{2}$   \\ 
$^{1}$Qualcomm Technologies, Inc., San Diego, California, U.S. \\
$^{2}$Automated Driving, QT Technologies Ireland Limited.}

\begin{document}
\maketitle
% \LK{We can adjust the figure sizes to make additional space as needed or place some side-by-side.  We still need conclusions text.}
\begin{abstract}
Autonomous driving systems require extensive data collection schemes to cover the diverse scenarios needed for building a robust and safe system. The data volumes are in the order of Exabytes and have to be stored for a long period of time (i.e., more than 10 years of the vehicle's life cycle). Lossless compression doesn't provide sufficient compression ratios, hence, lossy video compression has been explored. It is essential to prove that lossy video compression artifacts do not impact the performance of the perception algorithms. However, there is limited work in this area to provide a solid conclusion. In particular, there is no such work for fisheye cameras, which have high radial distortion and where compression may have higher artifacts. Fisheye cameras are commonly used in automotive systems for 3D object detection task. In this work, we provide the first analysis of the impact of standard video compression codecs on wide FOV fisheye camera images. We demonstrate that the achievable compression with negligible impact depends on the dataset and temporal prediction of the video codec.  We propose a radial distortion-aware zonal metric to evaluate the performance of artifacts in fisheye images. In addition, we present a novel method for estimating affine mode parameters of the latest VVC codec, and suggest some areas for improvement in video codecs for the application to fisheye imagery.

\end{abstract}    
\section{Introduction}
\label{sec:introduction}

In the recent years, autonomous vehicles are equipped with low-cost camera sensors that provide rich semantic information about the surrounding environment. In order to train robust deep learning algorithms that use camera data for perception tasks, training data is often collected across multiple vehicles and environmental conditions. This has led to a surge in camera data, and associated storage costs, which requires efficient and robust compression strategies. Autonomous driving systems also use  other sensors like Lidar but its volume is relatively small due to its sparsity \cite{mohapatra2021bevdetnet}.

Prior works~\cite{jpeg-comp,influence-avc-hevc,influence-night,kajak2020impact,friedland2020impact} have shown the impact of video coding standards such as AVC~\cite{wiegand2003overview} and HEVC~\cite{sullivan2012overview} on deep learning tasks. 
In~\cite{influence-avc-hevc}, the authors showed that HEVC and AVC data compression at Quantization Parameter (QP) less than 29 does not significantly affect the Faster R-CNN performance. Actually, they even showed that retraining the model with compressed data improved the Faster R-CNN model precision by 15\% compared to the model trained on uncompressed data. However, similar to the other prior works, their tests are limited to undistorted image compression and video input data.
Similarly, the authors in~\cite{Impact_DCNN} tested the impact of image compression across various deep learning tasks such as depth estimation, semantic segmentation, and showed that encoder-decoder architectures were more robust to extreme compression.

The authors~\cite{jpeg-comp} applied JPEG compression to the training data and showed negligible drop in performance while fine-tuning with the compressed data. In the case of real-world applications, it is important to train the model directly on compressed images with no prior knowledge about the uncompressed data, such that video compression techniques can be scaled for real-world storage applications. In another study that applied JPEG compression~\cite{friedland2020impact}, the authors reduced the input image complexity using JPEG, which resulted in similar accuracy with models trained using fewer parameters. Apart from object detection tasks, a recent study~\cite{tanaka2022does} evaluated the impact of compression on multi-object tracking accuracy (MOTA) against both Quantization Parameter and Motion Search Range (MSR). They showed significant impact on MOTA at 35 QP, while MSR did not have an impact on the performance.

Although most prior works compressed the image data using HEVC or AVC, the authors in~\cite{fischer2021robust} applied the VVC codec and showed that, at specific fine-tuning, the model's weighted average precision increased by 3.68\% compared to a model trained on uncompressed data. In addition, data augmentation with JPEG and VVC encoded images also resulted in improved weighted average precision. In case of night vision based pedestrian detection model~\cite{influence-night}, application of AVC compression to Far Infrared sensor data led to significant storage reduction. The AVC resulted in 0.5~Mbits/s data-rate for negligible loss in performance while JPEG resulted in 1~Mbits/s data-rate. The flexible macro-block segmentation tool of AVC helped in retaining the object details for improved performance, and generated lower data-rates compared to JPEG. 

The MPEG Video Coding for Machines (VCM) work~\cite{mpeg-vcm-use_cases, rozek2023video} differs from the current work in two fundamental ways. First, the data in VCM is all from camera without significant wide angle distortion. Second, the VCM work targets development of a codec with small impact on the visual tasks when the input is compressed.  The machine vision model used is developed presumably on uncompressed data. In a primary use case of training data storage, the role of compression is fundamentally altered. The data collected and used for training is compressed while the application of the trained model is generally on uncompressed data. Thus, the impact of compression on training data is of vital concern. Additionally due to the reversal of roles of compression, extremely high image quality may be desired in applying compressed data to the training process when application will be on uncompressed data. 

Most of the video compression techniques are tailored for human viewing and they are often applied only on undistorted images with a narrow FOV without modifying the underlying codec which is specifically designed for undistorted images. However, automotive camera suite has very wide angle cameras with high radial distortion due to the needs of a large horizontal field of view of 190$^{\circ}$ for near-field perception use cases. Four such wide angle fisheye cameras placed around the vehicle cover the full 360$^{\circ}$ field of view around the vehicle and form the basic sensor set in automotive systems for near-field sensing in combination with Ultrasonics sensors ~\cite{kumar2023surround, popperli2019capsule}. Relatively, fisheye camera perception has fewer literature as there are only a few public datasets. The limited available literature in various fisheye perception tasks such as object detection~\cite{rashed2020fisheyeyolo, yahiaoui2019overview}, semantic segmentation~\cite{sistu2019real,rashed2019optical}, depth estimation~\cite{kumar2021svdistnet, kumar2018near}, localization~\cite{tripathi2020trained}, soiling and weather detection~\cite{uricar2019desoiling, dhananjaya2021weather}, motion segmentation~\cite{mohamed2021monocular}, multi-task learning~\cite{sistu2019neurall, leang2020dynamic}, and near-field perception systems~\cite{eising2021near, dahal2019deeptrailerassist} indicate that special attention and radial distortion aware design is necessary.

To the best of our knowledge, given the lack of literature in understanding the impact of fisheye image compression on camera visual perception tasks, our main contributions are:
\begin{itemize}
    \item The impact of lossy compression of fisheye data on the object detection computer vision task is analyzed across various codecs. Our results show the highest compression that can be achieved without degrading the object detection performance on temporal and non-temporal datasets. 
    \item We emphasise on the necessity to apply lossy compression to the training data, and show the impact of fisheye compression on the object detection task while the model has no prior knowledge about the uncompressed dataset.
    \item Due to the high radial distortion in the image, unlike prior works that focused on full frame mAP to understand the impact of compression on undistorted images, we propose a radial distortion-aware zonal metric to analyze the impact of fisheye image compression. 
    \item Finally, we present a novel method to improve the existing VVC codec by adapting the camera motion model for the wide FOV camera (fisheye). 
\end{itemize}

Therefore, extending the work of Chan et al.~\cite{influence-avc-hevc} to wide FOV images, we are the first to apply the standard video compression codecs(HEVC, AVC) on wide FOV, fisheye images (\cite{fisheye8k, woodscape} to quantify both impact of lossy compression on Deep Learning model inference and training. Since standard compression codecs(HEVC, AVC) are designed for human visualization and undistorted images with the exception of VVC\cite{bross2021overview} that includes a general motion compensated prediction tools that can be applied to common wide FOV images, we present an improved motion model for VVC encoder using camera motion, intrinsics and extrinsics data.

\section{Video Compression of Wide FOV imagery}
\label{sec:video compression}

\begin{figure}
    \centering
    \captionsetup{font=small, belowskip=-8pt}
    \includegraphics[width=0.8\linewidth]{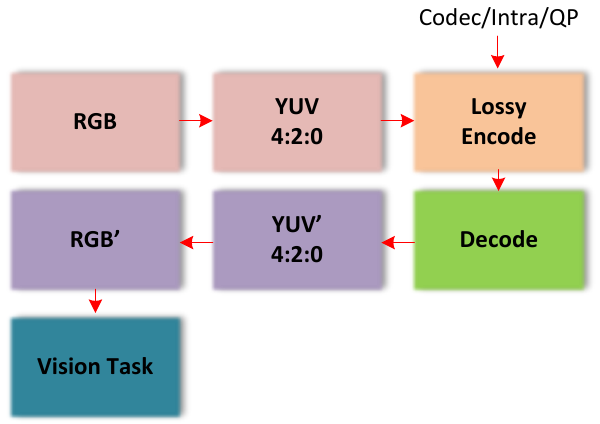}
    \caption{Video compression and evaluation system}
    \label{fig:video-system}
\end{figure}

%-------------------------------------------------------------------------
This work studies the use of standardized video codecs on wide FOV imagery. Video codecs utilize temporal prediction from one frame to another and can be greatly effective when the motion of the content matches the motion model of the video codec as the video encoder only needs to encode the motion compensated residual signal. In the case where the camera does not move, as in video surveillance, the background does not change, and hence, a video codec can successfully avoid repetition of unchanged data regardless of the complexity of the codec's motion model.
The major components of our evaluation are illustrated in Figure~\ref{fig:video-system}. First, original RGB images are converted to YUV color space and sub-sampled to 4:2:0 chroma format as this is efficiently handled by typical video codecs. Second, the images are provided to a lossy video encoder.  Several video encoder algorithms and fixed QP values are used to produce compressed bitstreams. Then, the bitstream is decoded and converted to modified YUV' and RGB' pixel values.  The RGB' images are provided as input to a Vision Task.  The Vision Task may be inference using an already trained model or may consist of training a model on the reconstructed images from the lossy codec. 

\subsection{Lossy Video Compression impact on inference mAP}
\label{sec:mAP inference imapct}
In order to understand the effect of the video compression artifacts on the inference mAP, it is necessary to evaluate the images on a standard deep learning network. Given the prevalence of object detection networks in the vision community and the performance vs. computation trade-offs offered by the single stage YOLO object detection network, we chose the YOLOv7 object detection network. We applied the AVC and HEVC codecs on the inference images and considered the set of images per camera as a video sequence and encoded the data at various Quantization Parameters (QPs).
The frames were compressed using ffmpeg and the decoded images were stored in png, lossless format. AVC and HEVC takes a minimum QP of 0 and a maximum QP of 51. A QP of 4 results in virtually lossless compression and QP of 51 leads to extreme compression. 

We test three different encoding methods using ffmpeg. ``HEVC intra'' uses the HEVC main profile but forces all frames to be intra-coded and hence no temporal prediction. ``HEVC main" directly uses the main profile of HEVC for encoding.  ``AVC" uses the main profile of AVC that allows temporal prediction but is a less capable codec than HEVC.  In all cases, ffmpeg defaults of other encoding parameters and complexity settings were used.

\subsection{Lossy Video Compression impact on training}
\label{sec:lossy compression training}
Typically, for most deep learning models, few hundred thousand frames of training data is required in order to achieve robust performance. By applying the codecs on inference data, the QP at which the mAP is least affected can be identified. This analysis can partially help in determining the ideal compression such that the inference performance is not affected. In real-world settings, it is not feasible to always store the original data to pre-train the model and then compress the data and use for further fine-tuning either. Therefore, we applied the compression on the training data at different QPs. A model was trained using compressed data (at each QP value) and the inference mAP was recorded on the original data. This analysis helps in identifying the ideal QP to use for compressing and storing only the compressed training data which can lead to valuable storage cost savings. 

\begin{figure}
    \centering
    \includegraphics[width=0.68\linewidth]{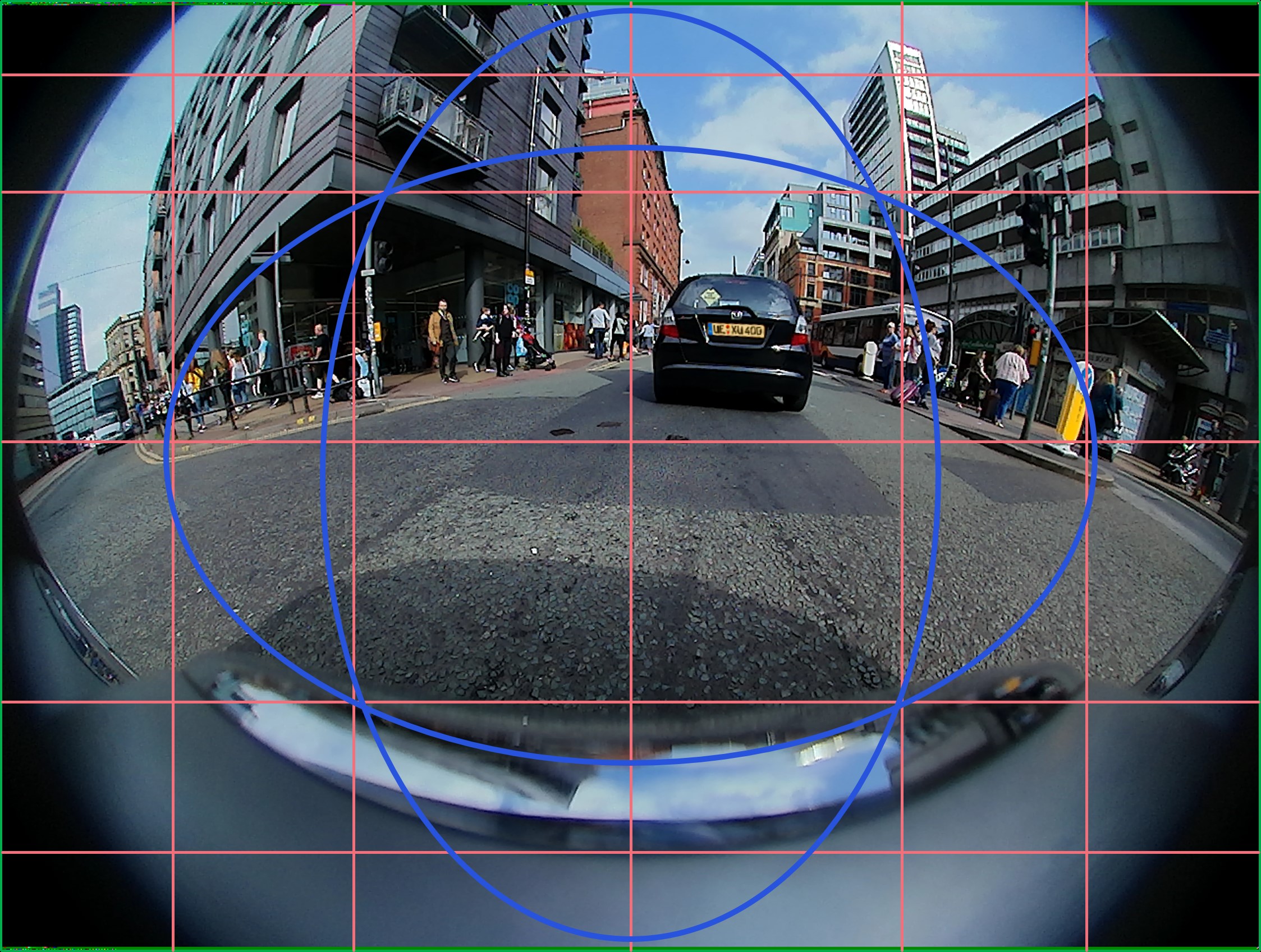}
    \caption{The area with the least distortion is defined as the union of the two ellipses and the objects inside this central region is evaluated in central mAP calculation while the objects outside this region is evaluated as peripheral mAP calculations.}
    \vspace{-0.3cm}
    \label{fig:fisheye_distortion}
\end{figure}

\textbf{Zonal Metric:} In \cite{fisheye_distortion}, the authors illustrated the fisheye distortion as a projection of an open cube using the 4\textsuperscript{th} degree radial polynomial distortion model. Therefore, in the fisheye images, squared grid becomes a curved box towards the periphery and motivate the need for curved bounding boxes.
Therefore, due to the high radial distortion at the periphery, it is important to understand the impact of fisheye image compression particularly at the periphery of an image. 
We define the zonal metric such that the objects are evaluated in either \textit{central mAP} calculation or \textit{peripheral mAP} calculation.
In the FOV of the camera images shown in Figure~\ref{fig:fisheye_distortion}, the straight lines parallel to the y-axis acts as the reference to indicate the curved nature of a straight building the image and similarly, the straight lines parallel to the y-axis shows the curvature of the windows which would otherwise be parallel to the y-axis in a pinhole camera image with no distortion. As the distortion increases towards the periphery of the image, we define the union of the two defined elliptical regions as the least distorted central region while the rest of the area is considered peripheral region. The elliptical regions depend on the particular radial distortion of the camera lens. To simplify calculations, this region could be approximated by a circle.

\subsection{Improved motion models}
Temporal prediction is very effective in reducing the redundancy in video compression where the camera has zero motion.
%with an extreme example seen with the FishEye8K data we used earlier (where the camera had zero motion).  
The effectiveness of temporal prediction depends on the underlying motion model of the codec.  In situations with significant camera motion, the ability to accurately represent the camera motion becomes important.  Traditional codecs use motion models based on 2-D block translation.  The recent VVC standard includes an affine mode defined by the motion of control points on the corners of a Coding Unit (CU). Careful analysis of these affine modes indicates the underlying motion is still 4x4 block translation.  In the VVC affine modes, a single 2D translation vector for each 4x4 block is determined using a 4-parameter or 6-parameter locally affine model, but the underlying model uses 4x4 block translation. The affine mode provides an efficient means of signaling a set of translation vectors in a CU composed of 4x4 blocks of pixels.

% \begin{figure}
%     \centering
%     \includegraphics[width=0.7\linewidth]{images/VVC_Affine.png}
%     \caption{VVC four parameter (2 control points) and six parameter (3 control points) motion models.}
%     \label{fig:VVC-Affine-Models}
% \end{figure}

In datasets with very low frame rate (i.e., where each frame in the scene is one or more seconds apart), motion models in temporal prediction are heavily limited, particularly when considering scale changes and camera lens radial distortion. Therefore, it is important to consider both the capability of the motion model and the practical aspect of selecting parameters for the model. 
A model with many parameters may provide an excellent motion prediction in theory but it may be impractical to estimate meaningful parameters in practice. 

\subsubsection{Epipolar geometry guided prediction}
\begin{figure}[t]
  \subfloat[WoodScape FV]{
  \begin{minipage}[c][1\width]{
	   0.23\textwidth}
	   \centering
	   \includegraphics[width=1\textwidth]{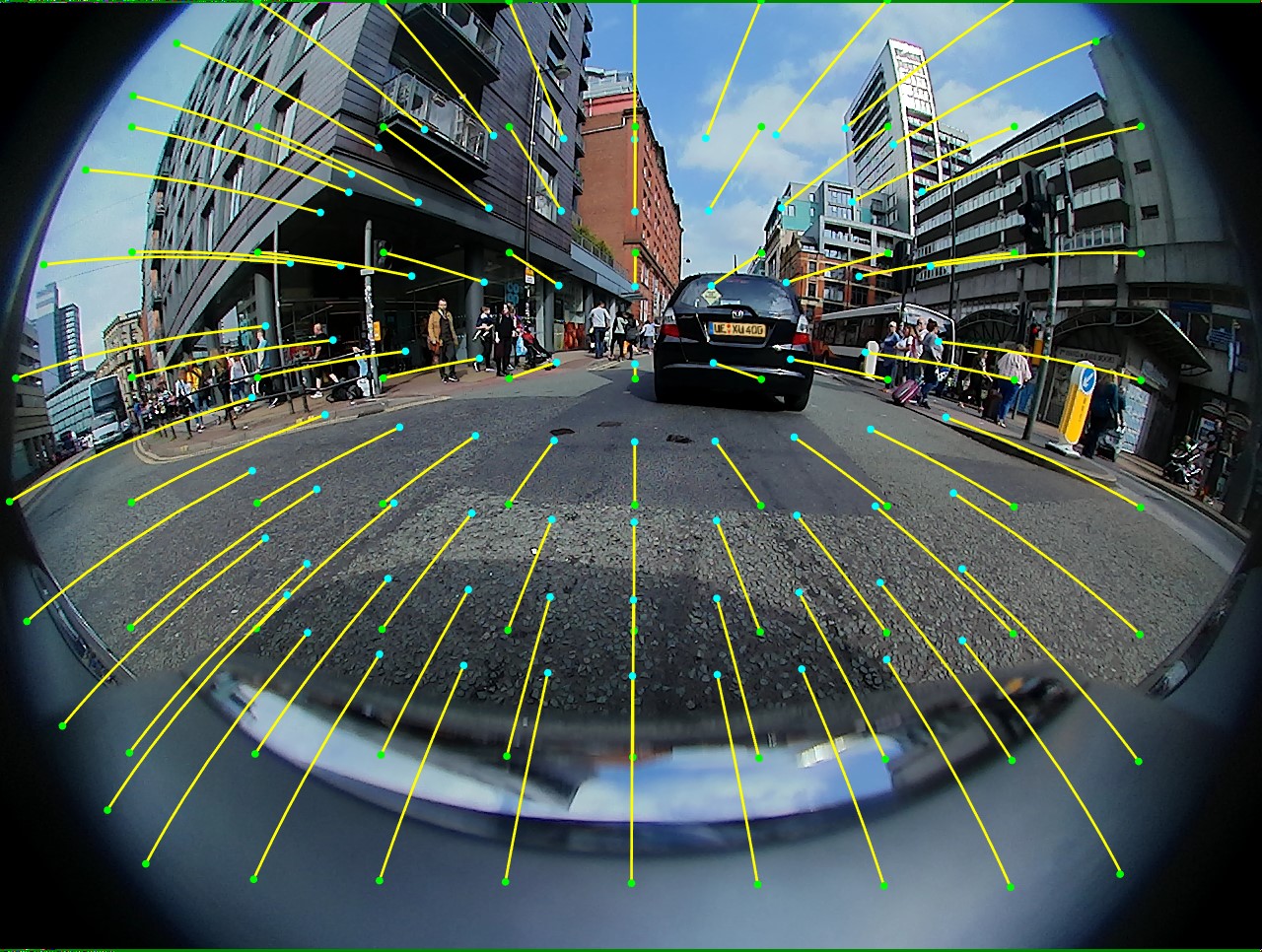}
	\end{minipage}}
 \hfill 	
  \subfloat[Woodscape MVR]{
	\begin{minipage}[c][1\width]{
	   0.23\textwidth}
	   \centering
	   \includegraphics[width=1\textwidth]{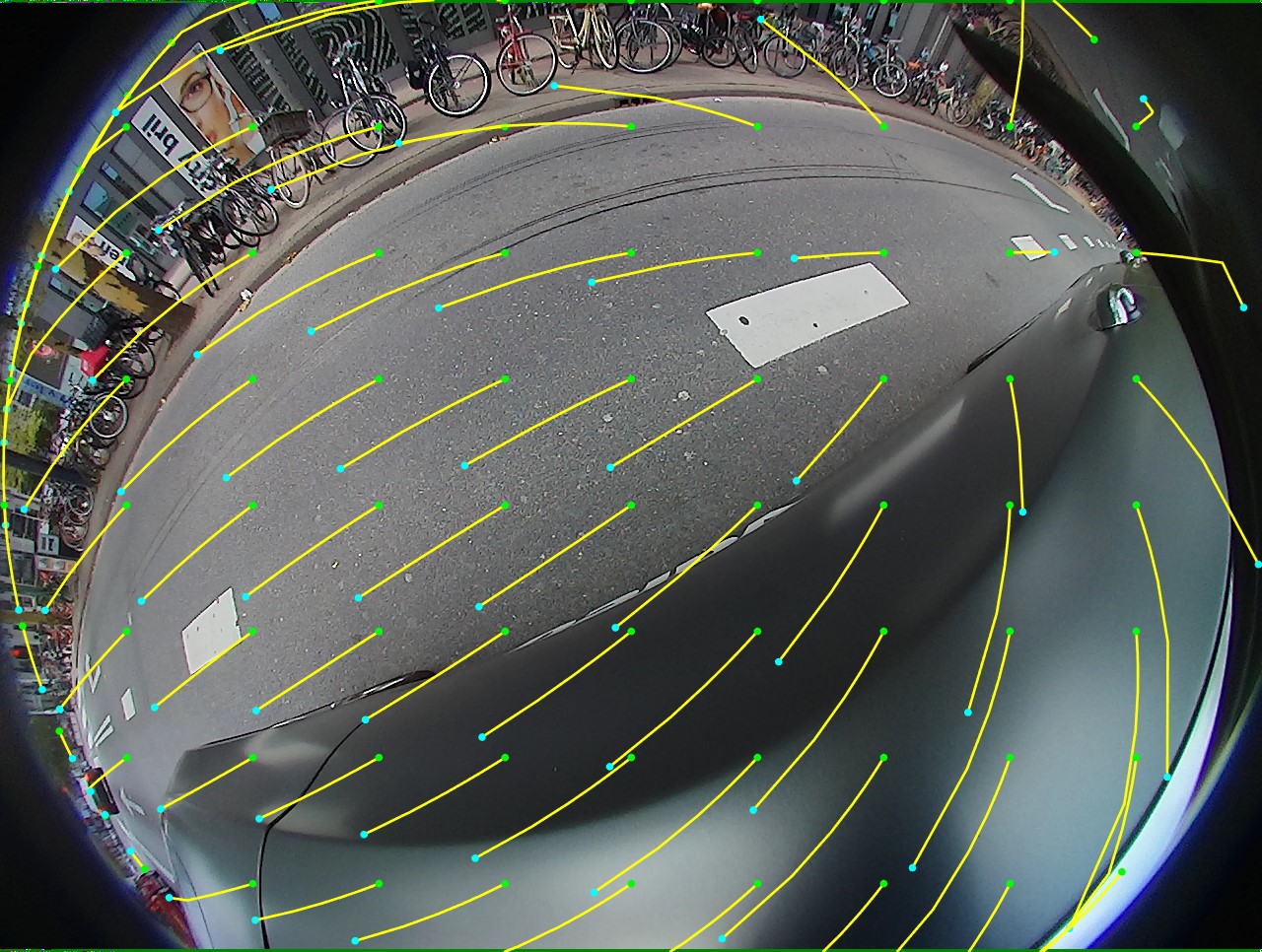}
	\end{minipage}}

\caption{Epipoles corresponding to ego motion of a camera position over time are shown for Woodscape FV (speed=6.2 m/s, yawrate = -0.6 degrees/sec, dt = 1s), and Woodscape MVR (speed=32.0 m/s, yawrate = 1.3 degrees/sec, dt = 1 s)}
\vspace{-0.3cm}
\label{fig:Epipoles}
\end{figure}

\begin{figure}[t]
  \subfloat[Reference ($t-1$)]{
  \begin{minipage}[c][1\width]{
	   0.23\textwidth}
	   \centering
	   \includegraphics[width=1\textwidth]{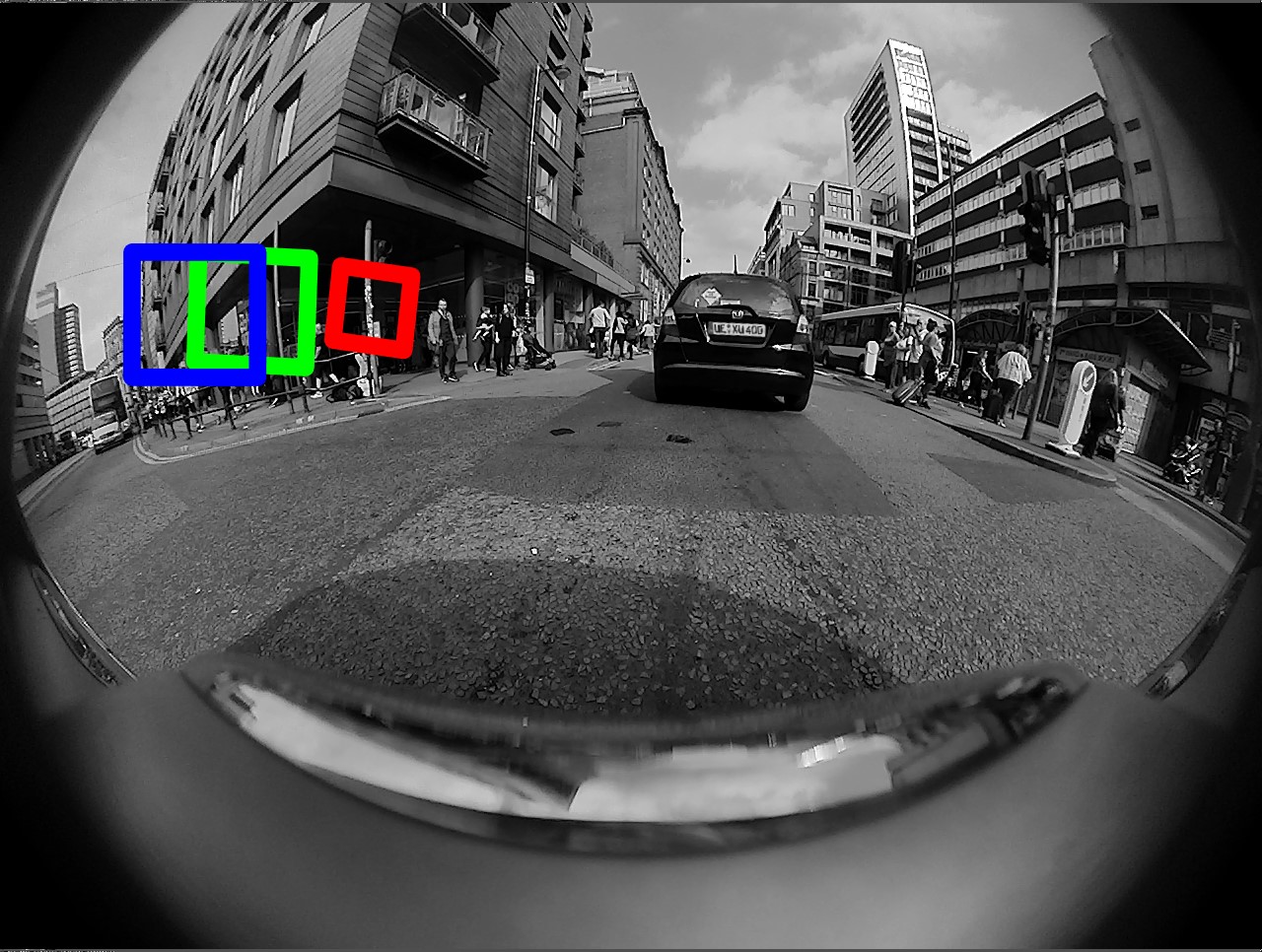}
	\end{minipage}}
 \hfill 	
  \subfloat[Target ($t$)]{
	\begin{minipage}[c][1\width]{
	   0.23\textwidth}
	   \centering
	   \includegraphics[width=1\textwidth]{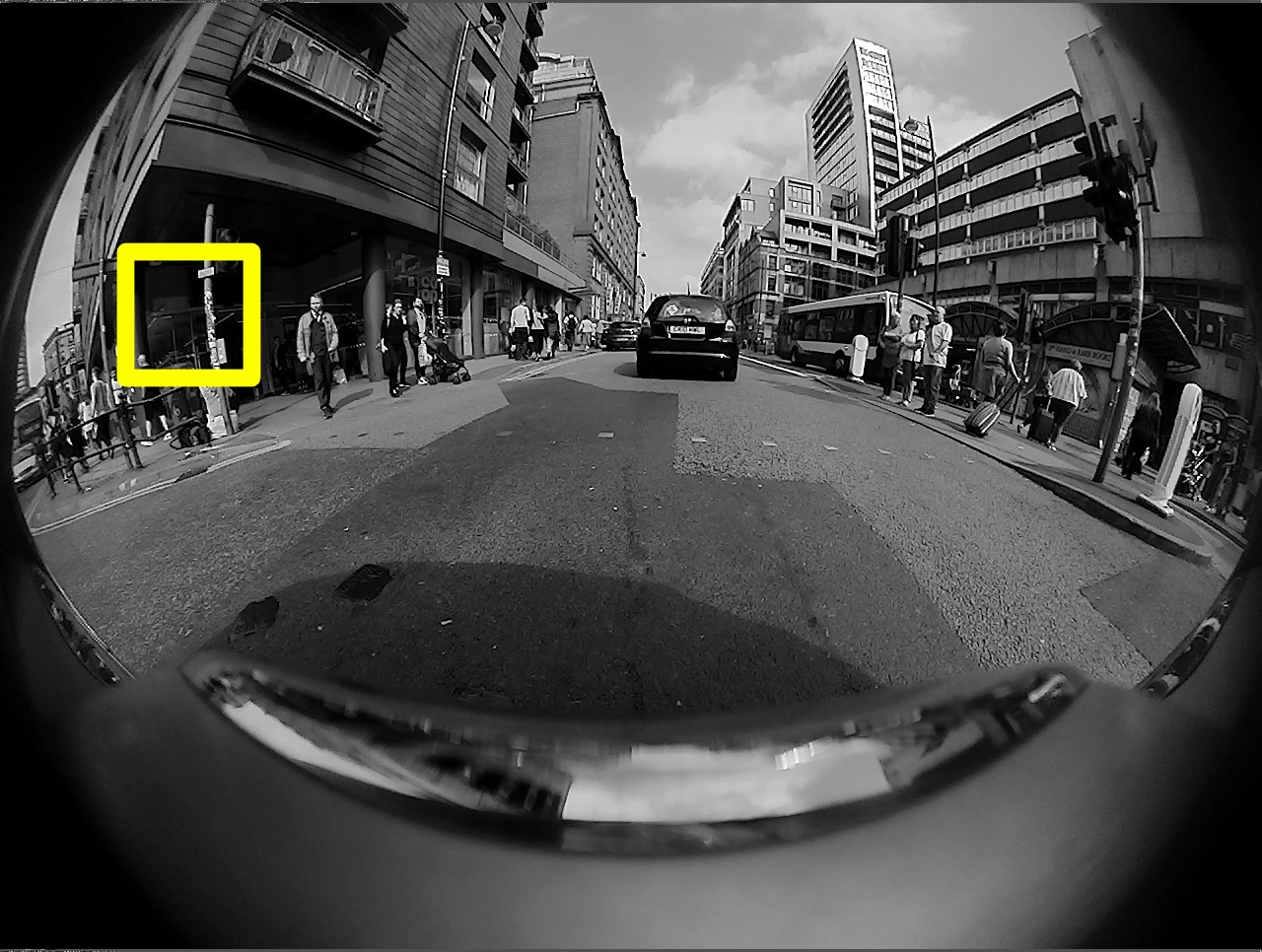}
	\end{minipage}}
\caption{Reference($t$) and Target($t+1$) images with overlay of blocks guided by epipole geometry.  Block size 128x128 is shown for visual clarity though smaller blocks sizes may be used.}
\vspace{-0.3cm}
\label{fig:Epipole Guided Predictors}
\end{figure}

Epipolar geometry relates to two overlapping camera views of a scene.  In our case, the two views are from the same camera but at different times where the camera has moved position.  Given the camera intrinsic, camera extrinsic and camera motion, it is possible to calculate the set of possible pixels in the first frame that correspond to a single pixel in the second frame. With a pinhole camera, this results in a line of possible positions. With a more general camera, the points will lie on a 1-D curve.  Examples of epipole curves of the WoodScape dataset are shown in Figure~\ref{fig:Epipoles} and matched blocks are illustrated in Figure~\ref{fig:Epipole Guided Predictors}.  The position on the curve depends on the depth in 3D of the point being imaged.  This assumes the scene is static between the two camera images.    

Knowledge of camera intrinsic, camera extrinsic, and camera motion greatly reduces the motion parameter estimation problem.  Consider a locally affine motion model defined on a block, the motion may be defined by control points, as in VVC with motion vectors (MV), at two or three corners of the block. The number of parameters creates a challenge for motion estimation. We propose to use the epipole geometry to greatly reduce the search space.  Given a reference and target frame along with the camera and motion information, we select a 1-D list of potential depth candidates.  The pixels corresponding to each corner of the block in the target frame at each candidate depth may be pre-computed. For each depth candidate, the corners of the block at give depth are mapped to a pixel in the reference frame.  Given a local block to predict, we form a predicted region on the reference frame by connecting the epipole locations of the corners at the candidate depth. 
Explicitly, the epipole geometry is used to predetermine the pixel domain displacement of each top corner grid point at a set of candidate depths giving a candidate MV at each grid point and depth $MV_{Epipole}[row][column][depth]$.  Given a 1-D list of $n+1$ candidate depth values $\{ d_0, d_1,... d_n\}$, $n+1$ candidate predictors are defined by the VVC motion model and motion vectors corresponding to the block corners in Equation~\ref{candidate-MV}.  The prediction of a block of pixels $P$ uses the VVC predictor given the current block location $(r,c)$, current block size, reference frame index $Idx$, and depth $d_i$, $P_i = VVC(r,c,size,Idx,mv_i[0],mv_i[1],mv_i[2])$.

\begin{equation} \label{candidate-MV}
\begin{aligned}
&mv_i[0] = MV_{Epipole}[row][column][d_i]\\
&mv_i[1] = MV_{Epipole}[row][column+1][d_i]\\
&mv_i[2] = MV_{Epipole}[row+1][column][d_i]\\
\end{aligned}
\end{equation}

\section{Experiments}
In order to evaluate the effect of lossy compression on deep learning models, we compressed the wide FOV images using AVC and HEVC codecs. The Woodscape fisheye camera~\cite{woodscape} images were chosen since this dataset consists of scenarios with both ego-vehicle motion and dynamic objects in the scene. Since the publicly available WoodScape dataset is sparsely sampled in time and does not show the true video nature of the camera images that are collected in real-world vehicles, we additionally applied HEVC and AVC compression to the FishEye8K surveillance camera images to evaluate the effectiveness of temporal prediction on the video data. 

\begin{figure*}
	\centering
	\begin{subfigure}{0.245\linewidth}
		\includegraphics[width=\linewidth]{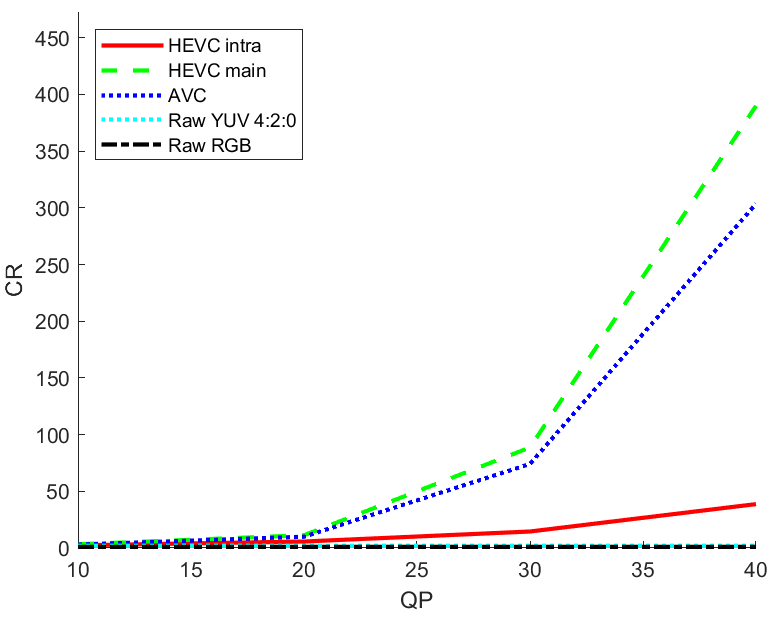}
		\caption{Fisheye8k}
		\label{fig:CR_fisheye8K_linear}
	\end{subfigure}
        \begin{subfigure}{0.245\linewidth}
		\includegraphics[width=\linewidth]{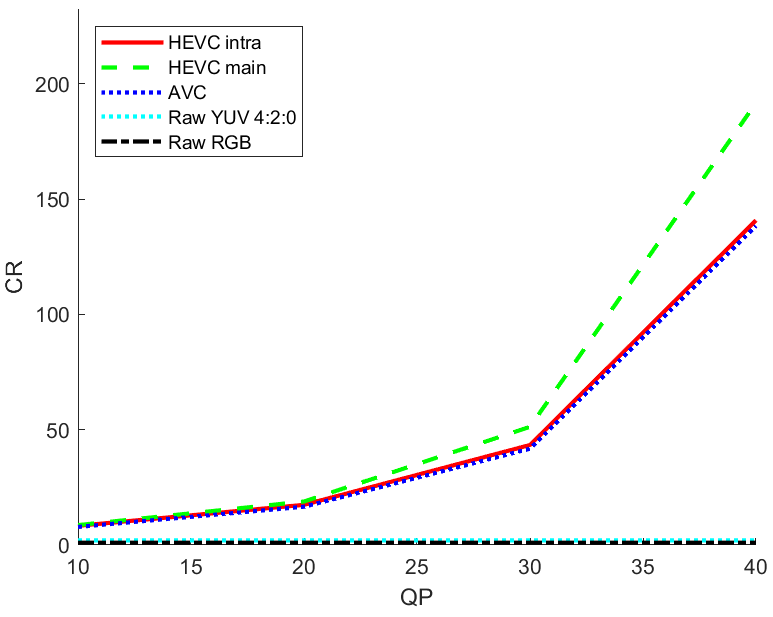}
		\caption{Woodscape}
		\label{fig:CR_Woodscape_linear}
	\end{subfigure}
        \begin{subfigure}{0.245\linewidth}
	    \includegraphics[width=\linewidth]{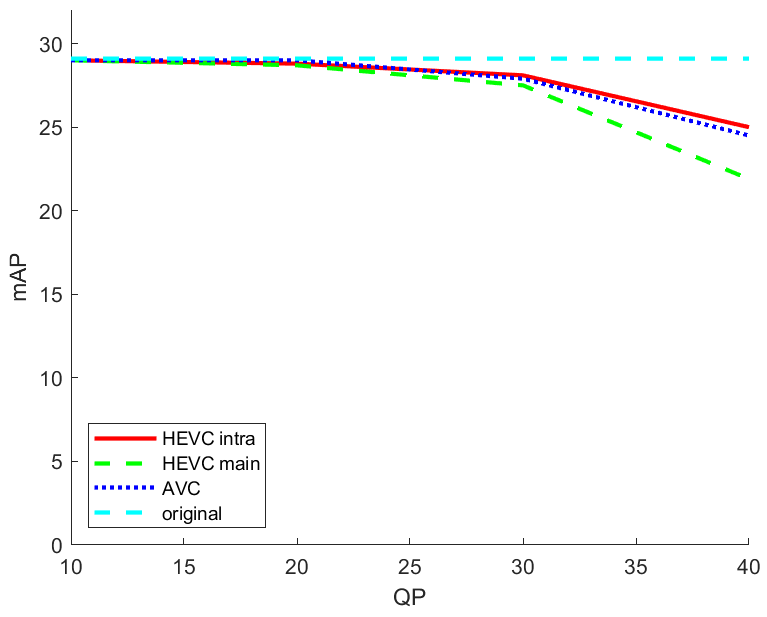}
	    \caption{Fisheye8k}
	    \label{fig:Woodscape mAP vs QP inference}
        \end{subfigure}
        \begin{subfigure}{0.245\linewidth}
	    \includegraphics[width=\linewidth]{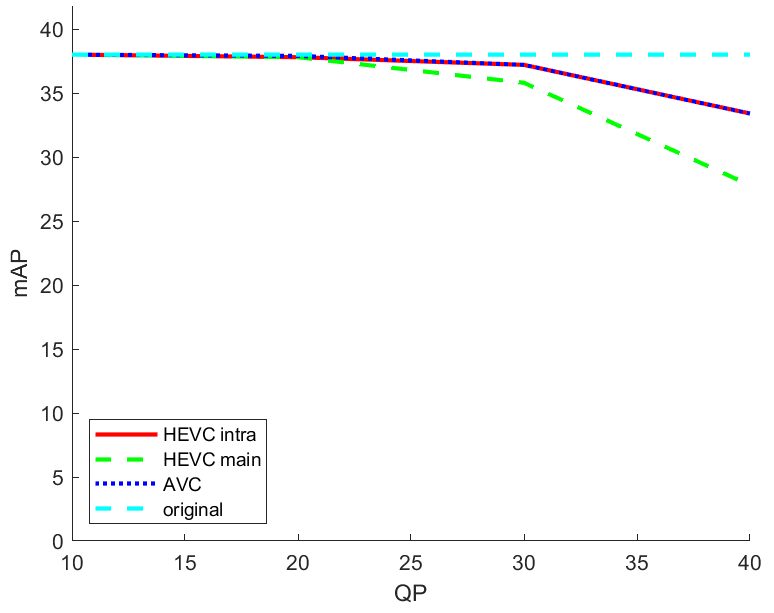}
	    \caption{Woodscape}
	    \label{fig:Fisheye8K mAP vs QP inference}
        \end{subfigure}
	\caption{(a) and (b): Compression ratio (CR) versus Quantization Parameter (QP) of a video codec applied to YUV 4:2:0 representation of content. (c) and (d): mAP versus QP of lossy video compression for various codecs.}
	\label{fig:subfigures}
\end{figure*}

\subsection{Dataset}
The WoodScape dataset~\cite{woodscape} consists of 10,000 fisheye camera images from 4 surround view cameras: front, rear, mirror left and mirror right. The data was collected across diverse geographical regions including USA, Europe and China. The authors provided 2D bounding box annotations for five classes: pedestrians, vehicles, bicycle, traffic lights and traffic signs. Since the dataset was released as a part of a challenge, the annotations are available only for the training set (8,234 images). Therefore, we split the training data into two sets of 5,762 images and 2,472 images for training and testing by maintaining an equal distribution of surround view cameras in both sets. 

The FishEye8K dataset~\cite{fisheye8k} was released with 22 videos (8,000 images) captured across 18 different fisheye cameras for traffic surveillance in Hsinshu, Taiwan. The authors provided 2D bounding box annotations for the entire dataset across Pedestrian, Bike, Car, Bus and Truck classes. Therefore, we used the author's training and validation split for training the YOLOv7 object detection model.

\subsection{YOLOv7 object detection model}
The YOLO (You Only Look Once) models are single stage object detection networks that predict both bounding boxes and classes. The Non-Maximal Suppression (NMS) post-processing is utilized to finalize the network's predictions. Compared to the latest YOLOR model, YOLOv7\cite{yolov7} achieves 0.4\% improved AP while reducing the computations by 15\% with 43\% fewer parameters. Notably, the authors improved the network's performance by improved model scaling and reparametrization planning techniques.

\subsection{Compression ratio results on inference data}
An example of the potential of 4:2:0 and video codec is illustrated in Figures~\ref{fig:CR_fisheye8K_linear},  
\ref{fig:CR_Woodscape_linear},  
by the compression ratio achievable on the FishEye8K surveillance dataset and the Woodscape automotive dataset. The chroma sub-sampling of the 4:2:0 format gives a 2:1 compression ratio compared to raw RGB data.  The image data is converted to YUV 4:2:0 color space and compressed with ffmpeg using different codecs and intra periods.  In all cases, the input consisted of a sequence of frames for a specific camera, and the results show the average compression ratio averaged across cameras for each codec tested at a specific Quantization Parameter (QP). On the FishEye8K data, prediction is effective as can be seen in comparing the HEVC intra (no motion compensation) with the HEVC main that utilises temporal prediction.  We see that even the AVC 420 codec, which includes temporal prediction, exceeds the HEVC codec when temporal prediction is removed by forcing all intra frames. Compression ratios over 50:1 can be achieved using video codecs at QP 30 using temporal prediction and the newer HEVC codec but, only about 43:1 using the HEVC-intra in case of the WoodScape dataset. 
%while eliminating the temporal prediction possibility from HEVC limits the compression ratio of HEVC intra to less than 40:1 in these tests. 
The motion model is not as effective on the Woodscape data due to the camera motion and the large temporal difference between frames over 1s even while compressing the same scene. However, the Fisheye8k dataset compression using the temporal prediction results in over 70:1 compression ratio while the all intra configuration results in only 14:1 compression ratio. Therefore, although the WoodScape data consists of repetitive patterns such as road and sky, the compression of the video data with complex Urban scene results in higher compression ratio assuming the motion model efficiently represents the motion in the content.

A central question is the impact of lossy video compression on DL tasks.  The first aspect of this evaluation is looking at lossy compression on images used for inference of a model trained without compression.  Different video codecs considered have different prediction modes, block size, spatial transforms, etc. (intra coding uses only spatial prediction within a single frame).  Despite these differences, the lossy quantization process is similar.
%though it may be applied to transform coefficients corresponding to different transforms or prediction error of the various codecs. 
We compare the performance of codecs across various QP parameter in Figures~\ref{fig:Woodscape mAP vs QP inference} and~\ref{fig:Fisheye8K mAP vs QP inference}.  We see that the difference in mAP between codecs is insignificant for \(QP < 20\). 

% \begin{figure}[ht]
% \centering
% \includegraphics[width=\columnwidth]{images/OmniCV_mAP-Fisheye8k_LogStats_Fisheye_2D_Object_detection_mAP_vs_QP_AdaptYmin_0.png}
% \caption{FishEye8K mAP versus QP of lossy video compression for various codecs}
% \label{fig:Fisheye8K mAP vs QP inference}
% \end{figure}

% \begin{figure}[ht]
% \centering
% \includegraphics[width=\columnwidth]{images/OmniCV_mAP-Woodscape_LogStats_Fisheye_2D_OD_mAP_vs_QP_AdaptYmin_0.png}
% \caption{Woodscape mAP versus QP of lossy video compression for various codecs}
% \label{fig:Woodscape mAP vs QP inference}
% \end{figure}

%The CR and mAP results versus QP can be expressed in a summary plots showing mAP versus CR. In figure \ref{fig:Fisheye8K mAP vs CR inference}, the benefit of temporal prediction of HEVC main and AVC can be seen in the high compression ratios achieved with small reduction in mAP.  When intra frame are forced with HEVC intra, so there is no temporal prediction, the CR is bounded below 40:1 even while the mAP reduction is nearly 5. In figure \ref{fig:Woodscape mAP vs CR inference}, the maximum compression ratio is lower and there is no visible impact of forced intra-coding.  This is presumably due to the limited temporal prediction effectiveness.

The mAP vs CR results for QP 30 on both Woodscape and FishEye8K are reported in Table~\ref{tab:CR_mAP}. The benefit of temporal prediction of HEVC-main and AVC can be seen in high compression ratios achieved with small reduction in mAP on the FishEye8K results. Even with a higher compression at QP 30, the drop in mAP is only around 1\% with HEVC-intra and AVC codecs. However, at and above QP 30, there is more than 2\% drop in mAP with HEVC-main (which may not be desirable for safety critical automotive applications) although HEVC-main consistently achieves better compression ratio then the other codecs/profiles. Therefore, as a trade-off, at QP20, the HEVC-main results in least drop in mAP for the best CR while, above QP30, both AVC and HEVC-intra results in improved trade-off between CR and mAP.  

For the Woodscape data, the HEVC main temporal prediction is less effective.  Comparing HEVC-Intra and HEVC-Main, the CR is increased slightly from 43.3 to 51.2 while the mAP is reduced from 37.2 to 35.8.  The AVC codec also includes temporal prediction and performs worst than the HEVC-Intra codec.  In both HEVC-Main and AVC, the motion model is limited to block translation which does not handle the fisheye radial distortion or zooming motion.  

\begin{table}[t]
\caption{\bf The mAP vs CR is captured across various codecs for QP 30 compressed data.}
\captionsetup{font=small, belowskip=-10pt}
\centering{
\begin{adjustbox}{width=0.7\columnwidth}
\small
\begin{tabular}{@{}clcc@{}}
\toprule
\cellcolor[HTML]{96bbce}\textit{Dataset}
& \multicolumn{1}{c}{\cellcolor[HTML]{fb9a99}\textit{Codec}} 
& \cellcolor[HTML]{fdbf6f}\textit{CR}   
& \multicolumn{1}{l}{\cellcolor[HTML]{00b050}\textit{mAP}} \\ 
\midrule
\multirow{4}{*}{Woodscape}  & Uncompressed              & 1    & 38.1 \\
                            & HEVC-Intra               & 43.3 & 37.2 \\
                            & HEVC-Main                & 51.2 & 35.8 \\
                            & AVC                       & 41.7 & 37.2 \\
\midrule
\multirow{4}{*}{Fisheye8K} & Uncompressed              & 1    & 29.1 \\
                            & HEVC-Intra               & 14.5 & 28.1 \\
                            & HEVC-Main                & 88.6 & 27.5 \\
                            & AVC                       & 74.1 & 27.9 \\ 
\bottomrule
\end{tabular}
\end{adjustbox}
}
\vspace{-0.3cm}
\label{tab:CR_mAP}
\end{table}
%------------------------------------------------------------
% \begin{center}
% \begin{table}[ht!]
% \centering
% \begin{tabular}{ |c|c|c|c| } 
% \hline
% Dataset & Codec & CR & mAP \\
% \hline
% \hline
% \multirow{3}{5em}{Woodscape} & Uncompressed & 1 & 38.1 \\
% & HEVC-intra & 43.3 & 37.2 \\ 
% & HEVC-main & 51.2 & 35.8 \\ 
% & AVC &  41.7 & 37.2\\
% \hline
% \multirow{3}{5em}{FishEye8K} & Uncompressed & 1 & 29.1 \\
% & HEVC-intra & 14.5 & 28.1 \\ 
% & HEVC-main & 88.6 & 27.5 \\ 
% & AVC &  74.1 & 27.9\\
% \hline
% \end{tabular}
% \caption{The mAP vs CR is captured across various codecs for QP30 compressed data.}
% \label{tab:CR_mAP}
% \end{table}
% \end{center}
%------------------------------------------------------------

%\MS{Any major difference between HEVC and AVC codec that can result in different types of artifacts?}
%\LK{I wrote some text suspecting human visual system optimizations in ffpeg and HEVC design}
% \begin{figure}[ht]
% \centering
% \includegraphics[width=\columnwidth]{images/OmniCV_mAP-Fisheye8k_mAP_vs_CR_AdaptYmin_0.png}
% \caption{FishEye8K: mAP versus CR}
% \label{fig:FishEye8K mAP vs CR inference}
% \end{figure}

% \begin{figure}[ht]
% \centering
% \includegraphics[width=\columnwidth]{images/OmniCV_mAP-Woodscape_mAP_vs_CR_AdaptYmin_0.png}
% \caption{Woodscape: mAP versus CR}
% \label{fig:Woodscape mAP vs CR inference}
% \end{figure}

\subsection{Lossy Video Compression impact on training}
\begin{figure}[ht]
\centering
\includegraphics[width=0.8\columnwidth]{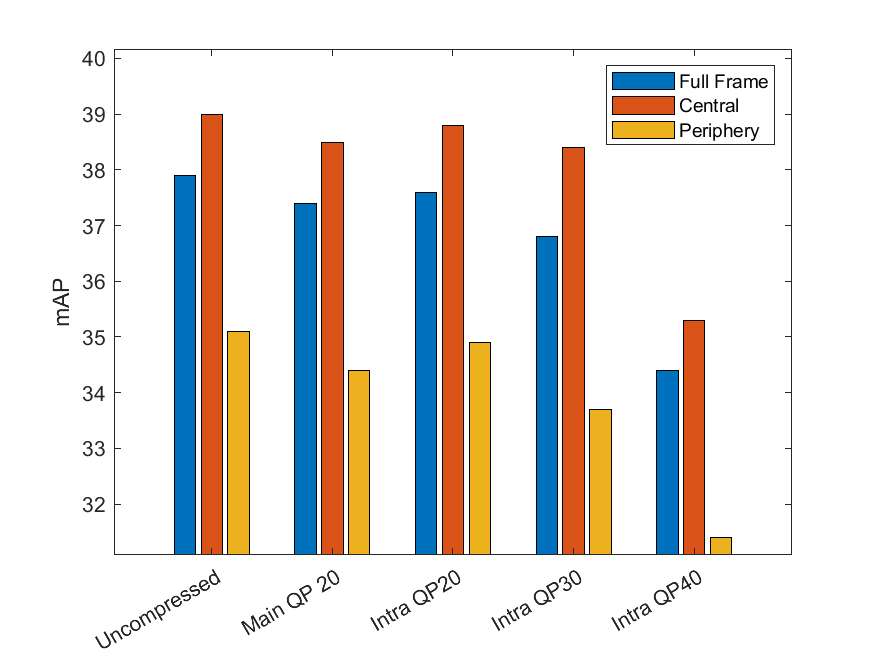}
\caption{The zonal mAP values for various models trained on the compressed data and evaluated on the uncompressed validation images.}
\label{fig:mAP retrained alternate}
\end{figure}

Figure~\ref{fig:mAP retrained alternate} shows uncompressed image inference mAP on full frame, central and periphery results on YOLOv7 models trained on training Woodscape images compressed using HEVC-main and intra profiles. The model trained and tested on uncompressed images is referenced as the baseline with 37.9\% mAP. 
The HEVC-intra QP 20 compressed trained model has a negligible drop in performance, and the models trained with higher QP compressed data results in lower mAP. 
%However, despite training the model on highly compressed QP 30 data, the overall inference drop compared to the baseline model is 1.1\%. 
The model trained on the original data and evaluated on the QP 40 compressed inference data results in 33.3\% mAP. However, the model trained on the QP 30 compressed model and QP 40 compressed inference data results in an increased 34.0\% mAP. Therefore, training on compressed data results in the model learning the compression artifacts and helps in recovering the model performance.  
%\vspace{-10pt}

To address the radial impact of radial distortion, we proposed the union of the elliptical regions as the central region with least distortion and the rest as the peripheral region. 
Across all the models, the central mAP is better than peripheral mAP. In case of the IntraQP20 model, the central and peripheral mAP almost retains the original uncompressed mAP performance. Therefore, the compression has minimal effect.  However at QP30, the central mAP drops only by 0.6\% compared to the original model's central mAP but the peripheral mAP drops by 1.4\%. Therefore, the compression has a more profound effect on the peripheral region at QP30 and overall mAP drop of 1.1\% at QP30 would not completely capture the effect of fisheye image compression and it is necessary to define zonal metric tailored to the camera parameters to identify the trade-off compression ratio.

\begin{figure}[t]
\centering
\includegraphics[width=0.8\columnwidth]{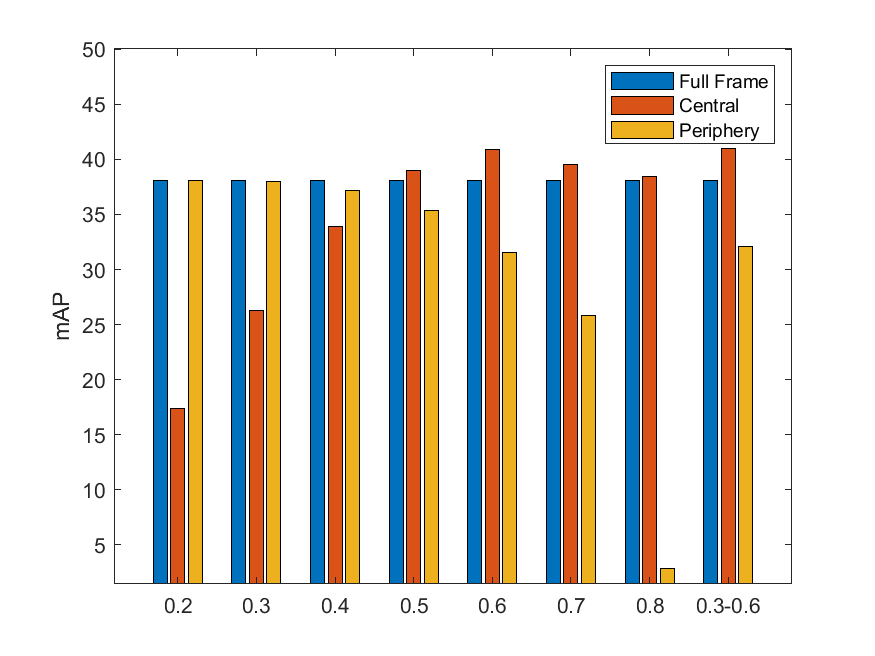}
\caption{The uncompressed model is evaluated on uncompressed validation images using zonal mAP for different radius values.}
\vspace{-0.3cm}
\label{fig:mAP retrained max_dist range}
\end{figure}
As shown in Figure~\ref{fig:mAP retrained max_dist range}, the central mAP increases as the distance increases (i.e., more objects are included in the central region). However, beyond a distance of 0.6, the peripheral mAP starts to drop due to higher distortion on the peripheral regions and poorer predictions. At the same time, the central mAP (at lower distances) with very few objects also has a lower mAP, possibly due to the stretched nature of the objects at the center of the image. The rightmost column in Figure~\ref{fig:mAP retrained max_dist range} only includes the objects between 0.3 and 0.6 and excludes both the stretched part of the image and the periphery. It shows higher mAP compared to other region definitions. Therefore, a more tailored zonal mAP that also considers the central objects stretching could be more informative to decide on the model's performance on the compressed data.

% \vspace{-10pt}
\begin{figure}[t]
\centering
\includegraphics[width=\columnwidth]{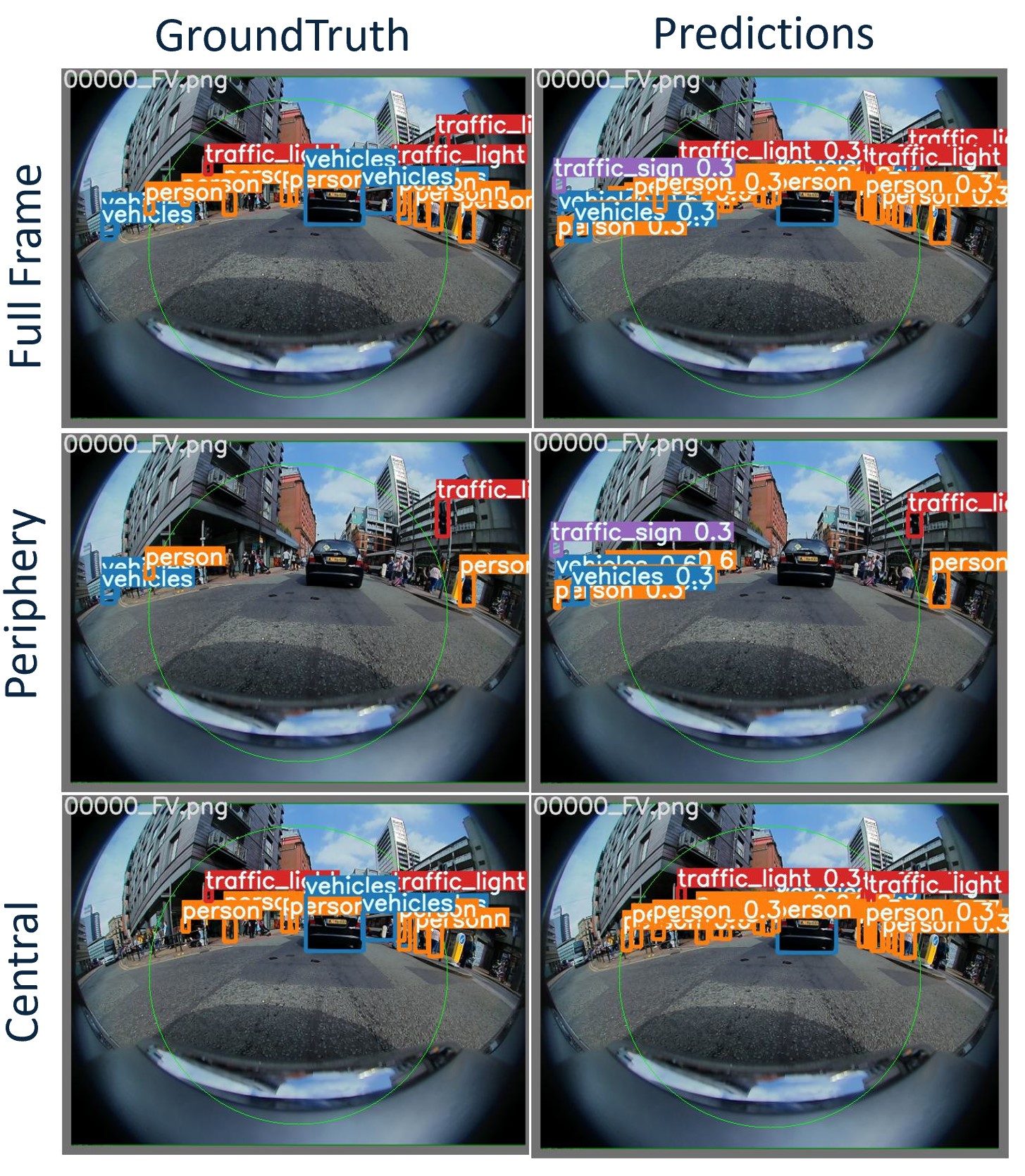}
\caption{The green circle defines the central regions and any object within the circle is evaluated as a part of central mAP and all the other objects are evaluated as a part of peripheral mAP.}
\vspace{-0.4cm}
\label{fig:mAP zonal}
\end{figure}

In Figure~\ref{fig:mAP zonal}, the objects in the scene are split based on the location into Central or Peripheral regions for the respective mAP calculation. The circle is defined from the center with a radius of 0.5 times the maximum distance (perpendicular distance) and the distortion of the objects at the periphery is significantly higher compared the central zone. Closer to the peripheral regions in the image, the objects become harder to detect due to distortion along with false positive detection such as the traffic sign prediction.

\section{Improving video codec motion models for wide FOV cameras}
The frame pairs from the Woodscape fisheye surround view camera images were selected and the epipole geometry was used to guide the target frame prediction based on the reference frame. The result of selecting the optimal value for this 1-D depth list at each block of the reference image provides a prediction image.
Table \ref{tab:Epipole_MSE} shows the average MSE result using the baseline zero motion model target image prediction and epipole guided motion model prediction of the target image. We tested on frame pairs from the WoodScape dataset across all the four surround view cameras. Due to the random sampling of the available Woodscape data, consecutive frame pairs were not readily available and most pairs had variable vehicle motion between frames. Therefore, given the above limitations and the lack of large scale video fisheye dataset with vehicle motion, we applied our proposed epipole guided search algorithm on the limited frame pairs in the dataset. However, on these challenging, complex Urban scenario images and across surround view cameras, our method resulted in 34.2\% MSE reduction while predicting the target image.

\begin{figure}[ht]
\begin{minipage}[b]{0.48\linewidth}
\centering
\includegraphics[width=\textwidth]{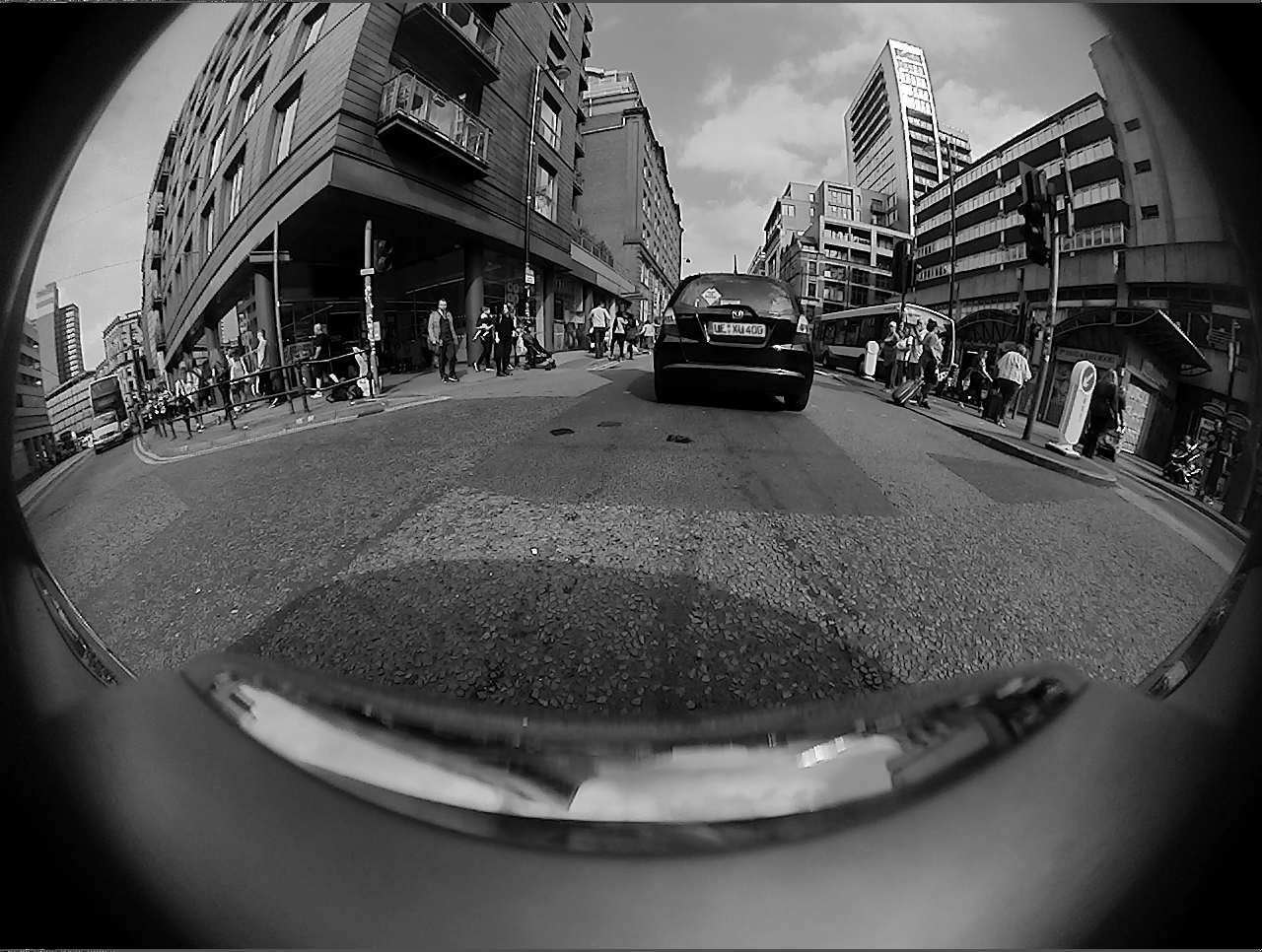}

\includegraphics[width=\textwidth]{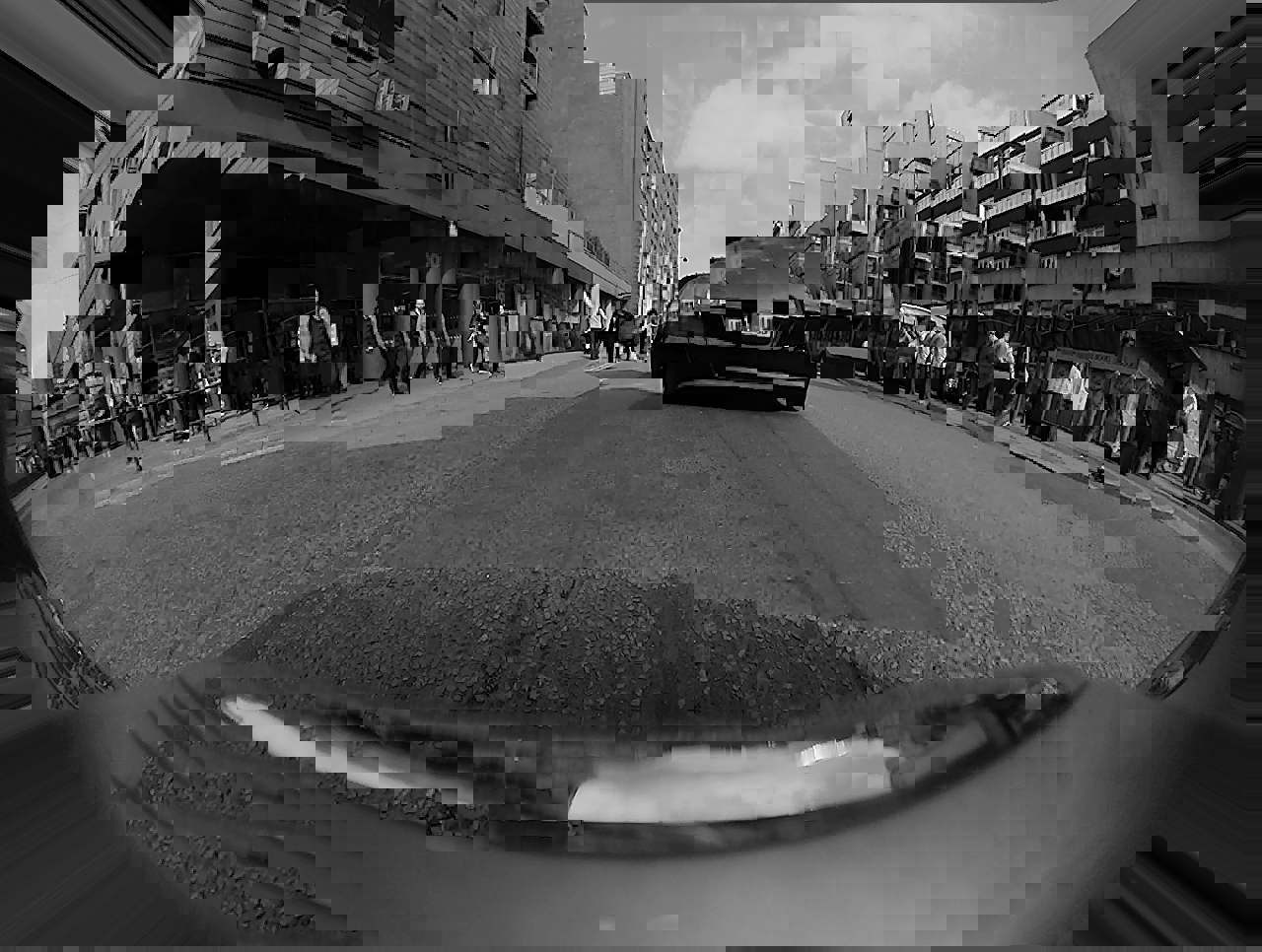}

\end{minipage}
\hfill
\begin{minipage}[b]{0.48\linewidth}
\centering
\includegraphics[width=\textwidth]{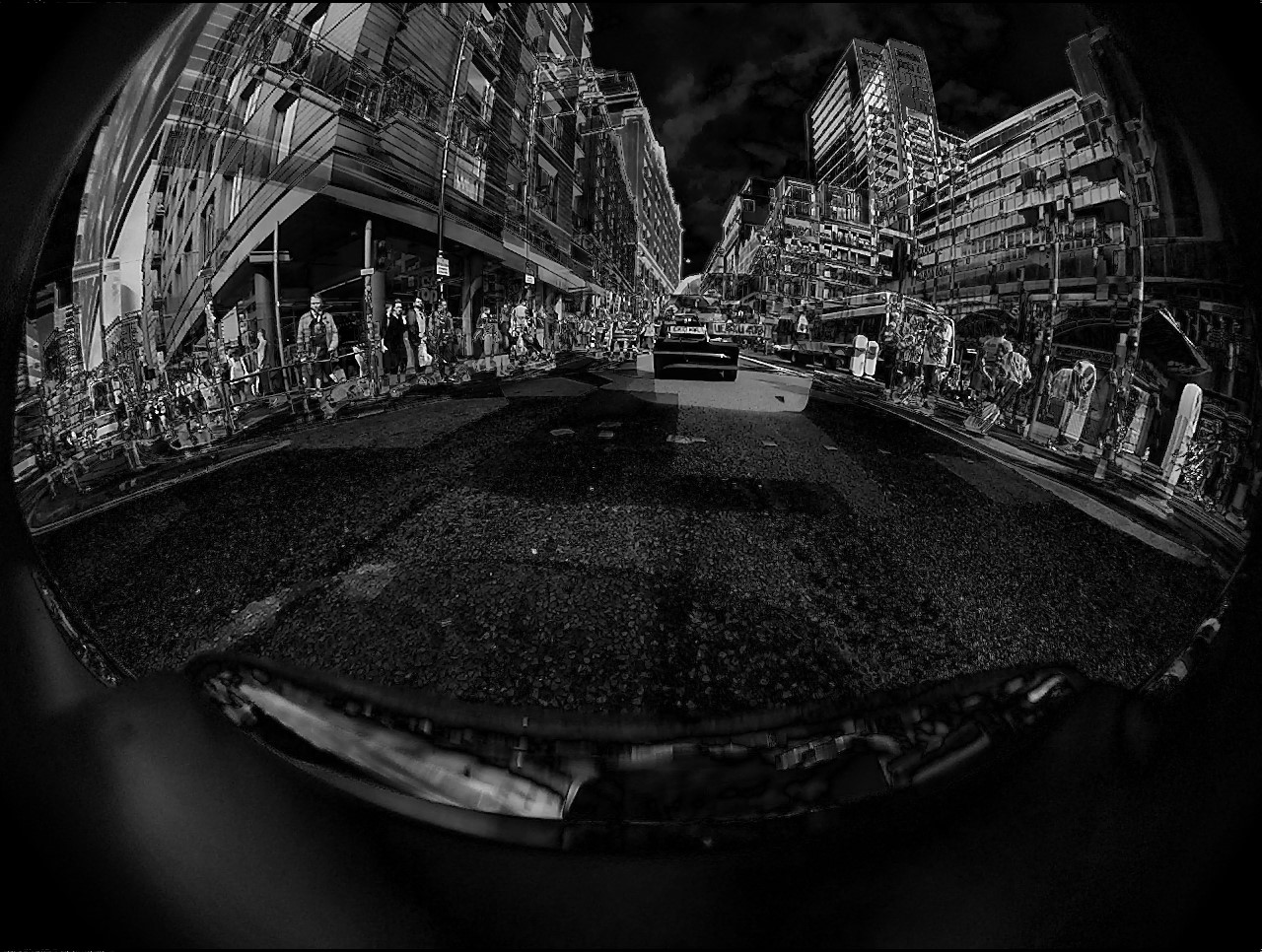}

\includegraphics[width=\textwidth]{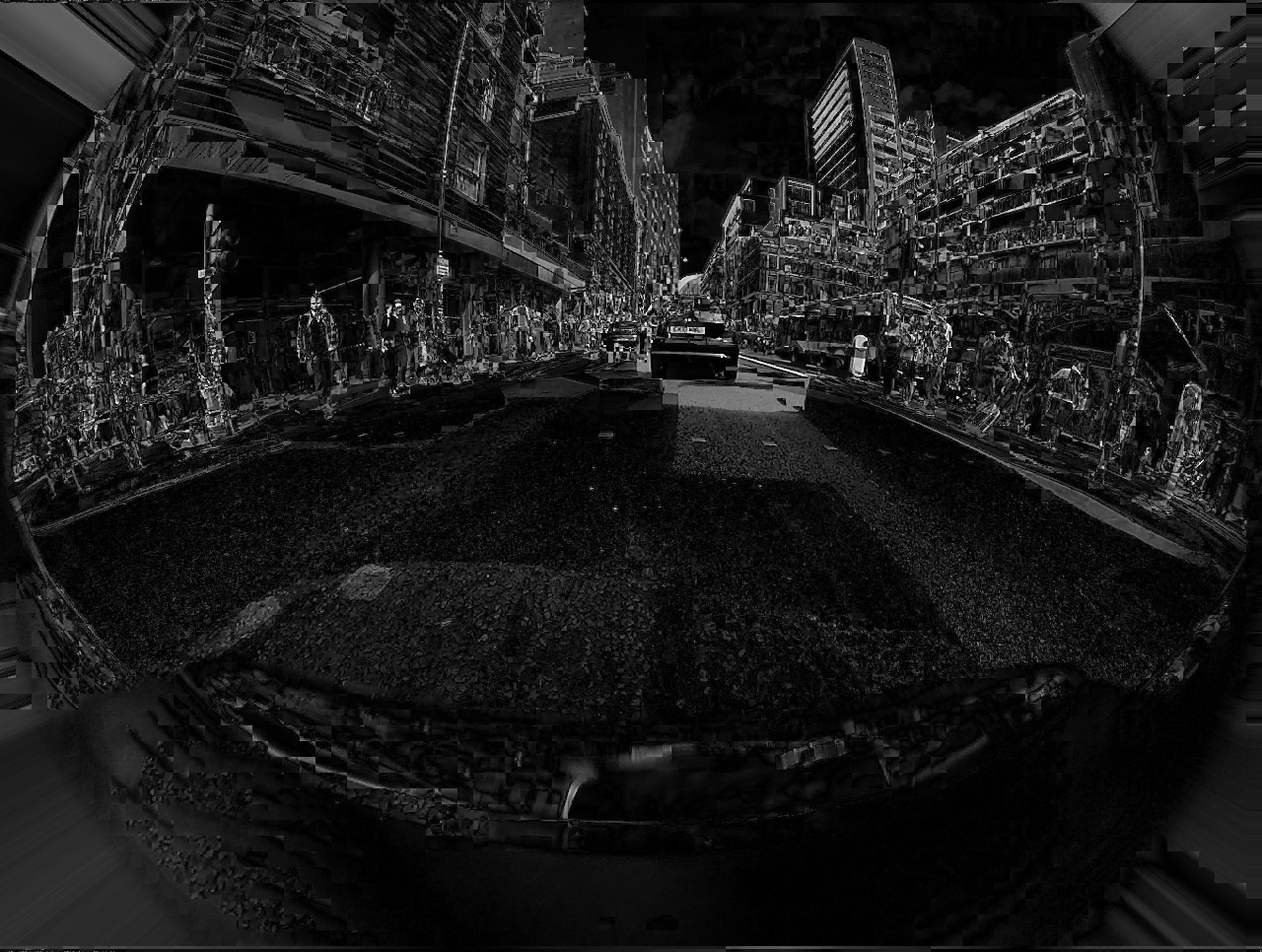}

\end{minipage}
\caption{WoodScape FV: Full prediction gray images and errors images corresponding to zero motion predictor and epipole guided locally affine predictor.  Top row zero motion predictor with error image (MSE 2939).  Second row, epipole guided 16x16 locally affine predictor and error image (MSE 1547).}
\vspace{-0.4cm}
\label{fig:VVC full frame prediction-FV}
\end{figure}

\begin{figure}[t]
\begin{minipage}[b]{0.48\linewidth}
\centering
\includegraphics[width=\textwidth]{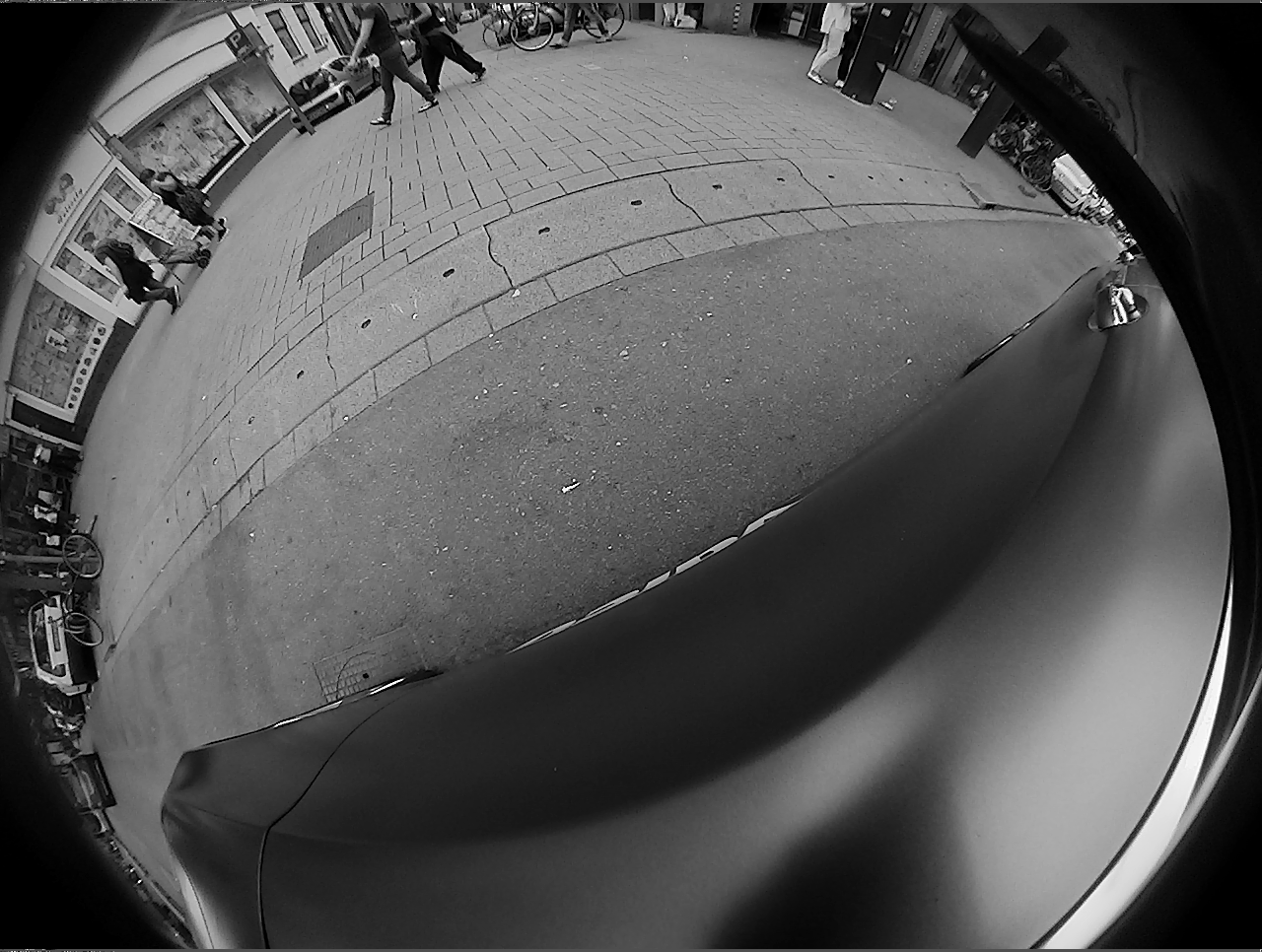}

\includegraphics[width=\textwidth]{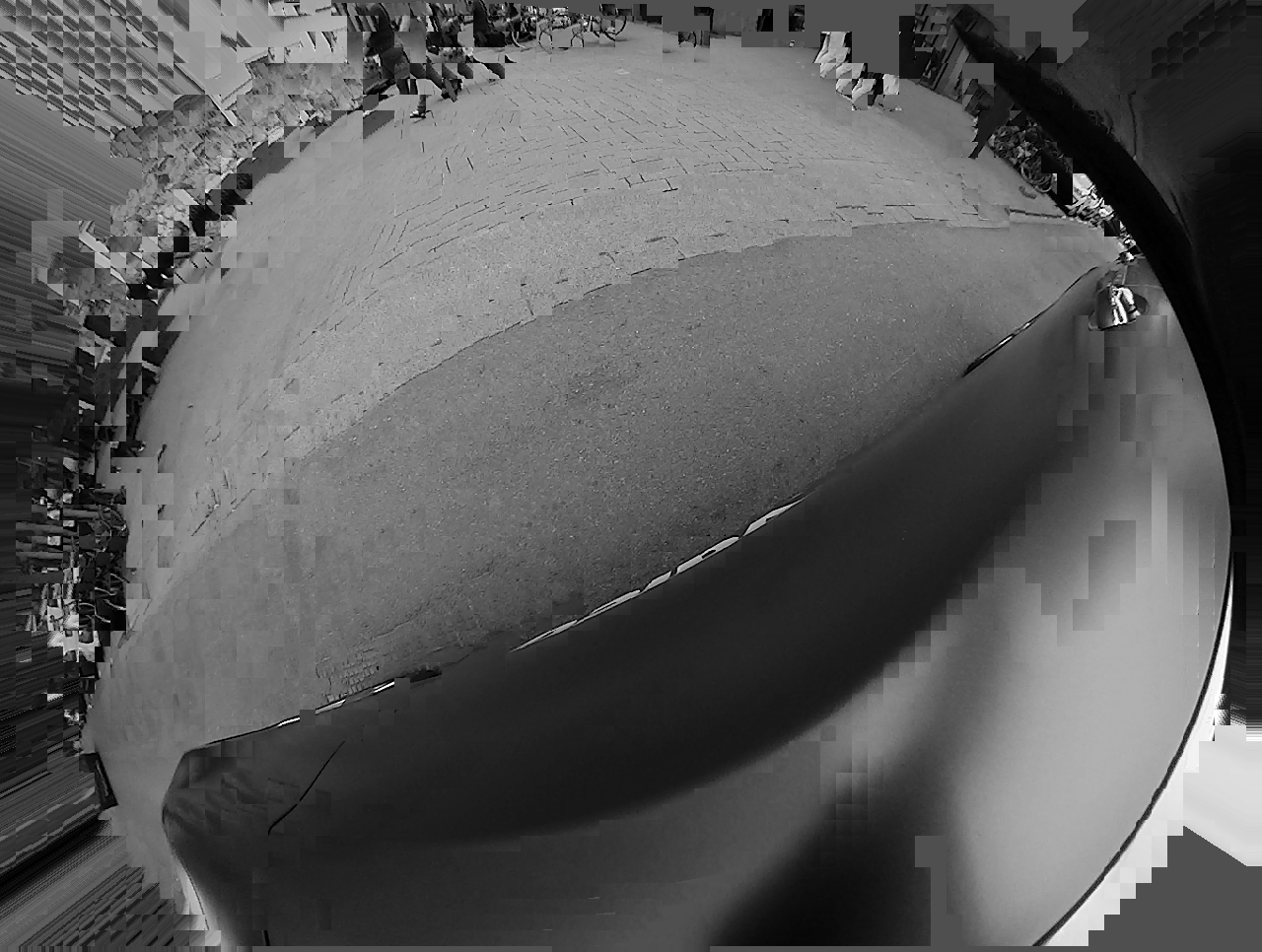}

\end{minipage}
\hfill
\begin{minipage}[b]{0.48\linewidth}
\centering
\includegraphics[width=\textwidth]{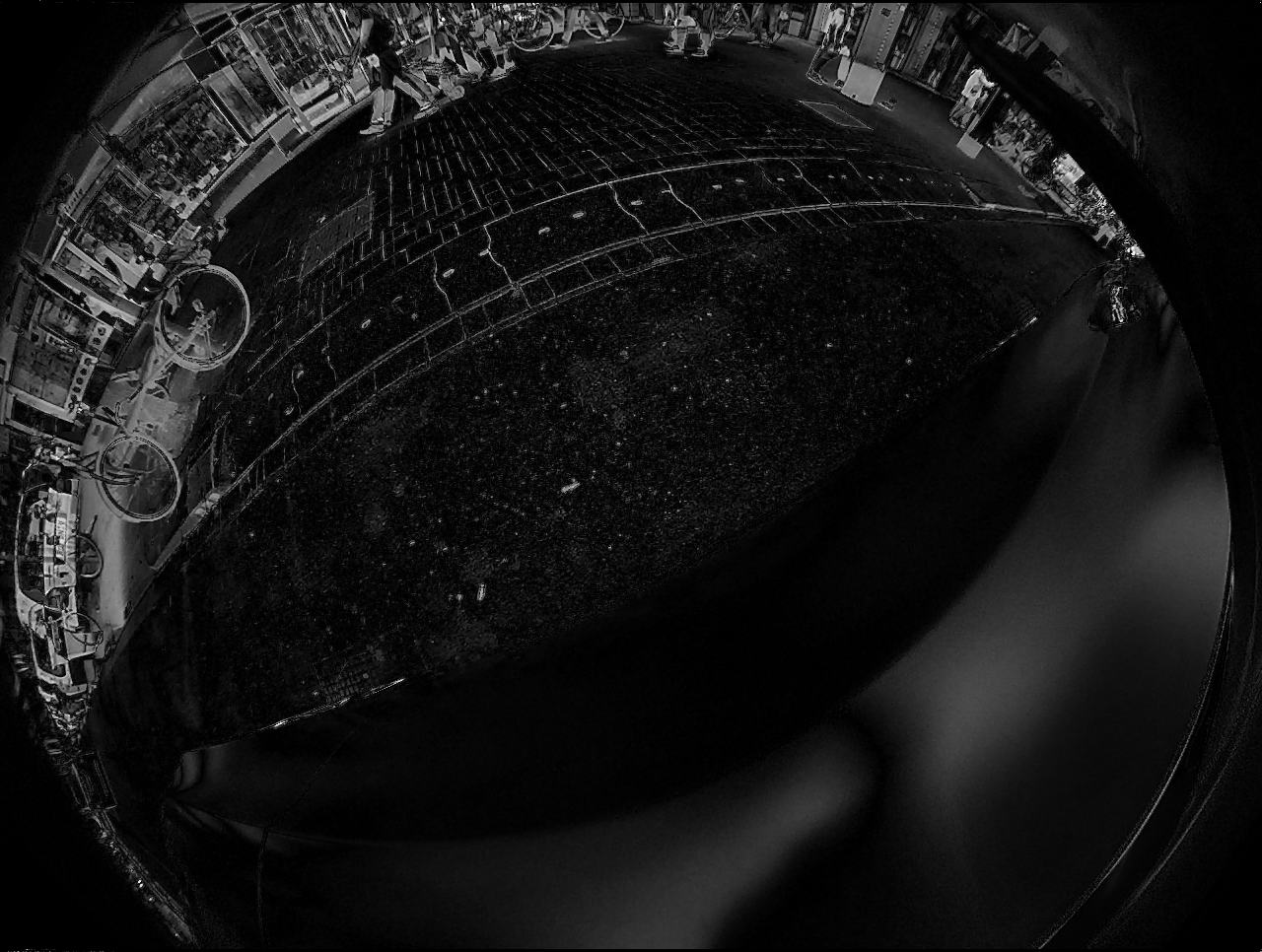}

\includegraphics[width=\textwidth]{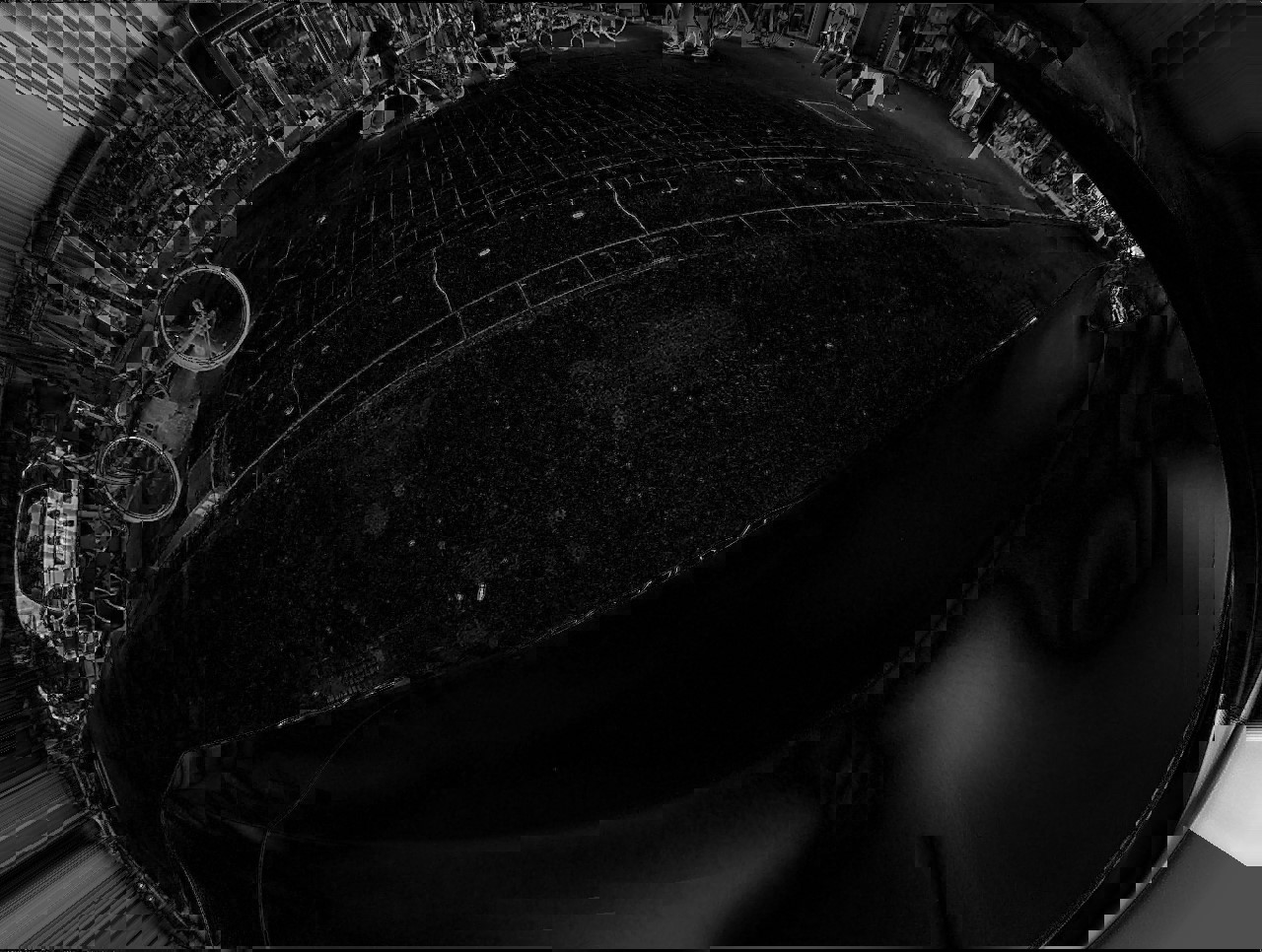}

\end{minipage}
\caption{WoodScape MVR: Full prediction gray images and errors images corresponding to zero motion predictor and epipole guided locally affine predictor.  Top row zero motion predictor with error image (MSE 1372).  Second row, epipole guided 16x16 locally affine predictor and error image (MSE 1269).}
\vspace{-0.4cm}
\label{fig:VVC full frame prediction-MVR}
\end{figure}

An example is shown in Figure~\ref{fig:VVC full frame prediction-FV}. The baseline prediction with a zero motion predictor on the front view image 1 results in 2929 MSE (Mean-Squared Error) while the epipole guided prediction results in 1547 MSE. In the second frame pair,
%\ref{fig:VVC full frame prediction-FV-frame4}, 
due to the lack of shadow and uniform texture of the road surface, the baseline zero motion MSE is 1910 while the epipole guided search results in 1372 MSE. Therefore, especially in case of front camera motion, the epipole guided prediction that takes camera intrinsic, extrinsic, and true motion results in improved prediction. Typically in the video codec, a lower MSE between the reference and the target frame will result in lower bit rate and hence improved compression ratio. In addition, we similarly tested on the mirror view right image (Figure~\ref{fig:VVC full frame prediction-MVR}) and compared to the baseline with an MSE of 1372. Our guided search results in an MSE of 1269. The MSE reduction with the MVR is less but the image is dominated by road surface and large time difference between the two frames which results in reduced temporal prediction effectiveness. 

% \begin{table}[h!]
% \centering
%  \begin{tabular}{|c |c |c|} 
%  \hline
%  Frame & Baseline & Epipole guided \\ [0.5ex] 
%  \hline\hline
%  FV1 & 2939 & 1547 \\
%  %\hline
%  FV2 & 1910 & 1263 \\
%  %\hline
%  MVR1 & 1372 & 1269 \\ %[1ex] 
%  \hline
%  \end{tabular}
%  \caption{The MSE between predicted image and target image using the baseline approach of zero motion model and Epipole guided search.}
% \label{tab:Epipole_MSE}
% \end{table}

% % %------------------------------------
% \begin{table}[!ht]
% \label{tab:Epipole_MSE}
% \caption{\bf The MSE between predicted image and target image using the baseline approach of zero motion model and epipole guided search.}
% \centering
% \setlength{\tabcolsep}{0.3em}
% \begin{tabular}{@{}lcc@{}}
% \toprule
% \multicolumn{1}{c}{\cellcolor[HTML]{96bbce}\textit{Frame}} 
% & \cellcolor[HTML]{fb9a99} \textit{Baseline} 
% & \cellcolor[HTML]{00b050} \textit{Epipole guided} \\ 
% \midrule
% FV1                                & 2939 & 1547 \\
% FV2                                & 1910 & 1263 \\
% MVR1                               & 1372 & 1269 \\ 
% MVL1                               & 2883 & 1727 \\
% RV1                                & 6053 & 4166 \\
% \bottomrule
% \end{tabular}
% \end{table}
% %------------------------------------

% %------------------------------------
\begin{table}[t]
\caption{\bf The MSE between predicted image and target image using the baseline approach of zero motion model and epipole guided search across camera views.}
\centering
\setlength{\tabcolsep}{0.3em}
\begin{tabular}{@{}lccc@{}}
\toprule
\multicolumn{1}{c}{\cellcolor[HTML]{96bbce}\textit{Frame}} 
& \cellcolor[HTML]{fb9a99} \textit{Baseline} 
& \cellcolor[HTML]{E59866} \textit{Epipole guided}
& \cellcolor[HTML]{00b050} \textit{MSE change[\%]} \\ 
\midrule
WoodScape                     & 3031.4 & 1994.4 & 34.2\%\\
\bottomrule
\end{tabular}
\label{tab:Epipole_MSE}
\end{table}
%------------------------------------

% \begin{figure}[ht]
% \begin{minipage}[b]{0.48\linewidth}
% \centering
% \includegraphics[width=\textwidth]{images/Frame_4_GrayReference_WoodScape_FV_to_WoodScape_FV.png}

% \includegraphics[width=\textwidth]{images/Frame_4_Predicted_WoodScape_FV_to_WoodScape_FV_AffineSix_16_[16x16]_W0_SubBlockSize_1.png}

% \end{minipage}
% \hfill
% \begin{minipage}[b]{0.48\linewidth}
% \centering
% \includegraphics[width=\textwidth]{images/Frame_4_ErrorImageZeroMotion_WoodScape_FV_to_WoodScape_FV_MSE_1910.png}

% \includegraphics[width=\textwidth]{images/Frame_4_ErrorImage_WoodScape_FV_to_WoodScape_FV_AffineSix_16_[16x16]_W0_SubBlockSize_1_MSE_1263.png}

% \end{minipage}
% \caption{Woodscape FV: Full prediction gray images and errors images corresponding to zero motion predictor and epipole guided locally affine predictor.  Top row zero motion predictor with error image (MSE 1910).  Second row, epipole guided 16x16 locally affine predictor and error image (MSE 1263)}
% \label{fig:VVC full frame prediction-FV-frame4}
% \end{figure}

\subsection{Suggestions to improve motion models used in future video codecs}

To support cameras with significant lens distortion and camera motion, we need to improve the motion model in order to have accurate temporal prediction.  We suggest the following:
%Initially, the locally affine model of VVC was of interest but detailed study indicated that these modes fundamentally rely on traditional block translation. 
\begin{itemize}
    \item \textbf{True local affine model}: The local adaptivity should be able to support spatially varying camera lens distortion but a true local affine model appears desirable to improve temporal prediction and hence compression.
    \item \textbf{Efficient signaling} of the improved model is desired to avoid overhead in transmitting the motion information. The current VVC syntax for signaling affine mode requires signaling several parameters and multiple motion vectors at each CU and is not expected to be efficient when motion is dominated by affine elements with different parameters such as a zoom due to fast camera motion of a vehicle.  Thus, an efficient means of signaling large regions of affine motion is an anticipated need.
    \item \textbf{Epipole guided search}: An efficient means for determining motion parameters, though not officially part of the standard, is essential for the improved motion tool to be useful in practice. The epipole guided search can reduce a 4 or 6 parameter motion search to searching a 1-D list of candidates distances.
    \item \textbf{Dataset}: An important step to move forward with producing a codec for these applications is for the community to develop a dataset with true video motion (15 fps or 30 fps), unlike the WoodScape dataset, and significant camera motion, unlike the FishEye8K dataset.
\end{itemize}

\section{Conclusion}
We presented the study on the impact of lossy fisheye video compression on a camera visual perception task i.e., 2D fisheye object detection. Due to the excessive storage costs involved in saving the automotive driving data, it is important to first understand the extent to which the fisheye data could be compressed without affecting the end task. Our results showed that a minimum of 10x compression ratio is achievable for a negligible drop in mAP and over 80x compression ratio is achievable for a 1-2\% drop in mAP for static camera sequences.
Although the overall mAP is informative in deciding the ideal compression ratio for pinhole camera models, due to the high distortion at the periphery a custom metric per camera model is required. Therefore, we present a novel zonal mAP metric to highlight the effect of compression artifacts on the high distortion regions in the image which ensures that the optimal compression ratio is chosen while the adverse effects of compression artifacts on the performance of the model are avoided. 
Finally, since video compression achieves remarkable compression rates by exploiting temporal correlations between successive video frames, an accurate motion model is mandatory. The existing codecs rely on a block translation motion models which give sub-optimal temporal prediction with high-speed camera motion and wide-angle camera distortion. Therefore, we present an epipole guided motion prediction model which results in 34\% lesser MSE compared to the baseline which translated to lower bitrate requirement for storing and transmitting the compressed data. 
In future work work, we plan to investigate the impact on a larger set of vision tasks, including temporal tasks such as tracking develop improved motion models to include in future video codecs based on wide FOV and high-speed camera motion while dealing with dynamic objects in the scene.

{
    \small
    \bibliographystyle{ieeenat_fullname}
    \bibliography{main}
}

% WARNING: do not forget to delete the supplementary pages from your submission 
% \input{sec/X_suppl}

\end{document}